\documentclass[11pt,a4paper]{article}
\usepackage{amssymb}
\usepackage{amsmath}
\usepackage{subfig}
\usepackage{multirow}
\usepackage{bm}
\usepackage{mathtools}
\usepackage[hyperref]{naaclhlt2019}
\usepackage{times}
\usepackage{latexsym}
\usepackage{enumitem}
\usepackage{colortbl}
\usepackage{booktabs}
\usepackage{arydshln}
\usepackage[normalem]{ulem}

\usepackage{url}

\aclfinalcopy 
\def\aclpaperid{1617} 

\title{Argument Mining for Understanding Peer Reviews}

\author{Xinyu Hua, Mitko Nikolov, Nikhil Badugu, Lu Wang\\
  Khoury College of Computer Sciences\\ Northeastern University \\ Boston, MA 02115 \\
  {\tt \{hua.x, nikolov.m, badugu.n\}@husky.neu.edu}\\
  {\tt luwang@ccs.neu.edu} 
 }

\begin{document}
\maketitle
\begin{abstract}
Peer-review plays a critical role in the scientific writing and publication ecosystem. To assess the efficiency and efficacy of the reviewing process, one essential element is to understand and evaluate the reviews themselves. In this work, we study the content and structure of peer reviews under the argument mining framework, through automatically detecting (1) argumentative propositions put forward by reviewers, and (2) their types (e.g., evaluating the work or making suggestions for improvement). 
We first collect $14.2$K reviews from major machine learning and natural language processing venues. $400$ reviews are annotated with $10,386$ propositions and corresponding types of \textsc{Evaluation}, \textsc{Request}, \textsc{Fact}, \textsc{Reference}, or \textsc{Quote}. 
We then train state-of-the-art proposition segmentation and classification models on the data to evaluate their utilities and identify new challenges for this new domain, motivating future directions for argument mining. 
Further experiments show that proposition usage varies across venues in amount, type, and topic.
\end{abstract}

\section{Introduction}
\label{sec:intro}
Peer review is a process where domain experts scrutinize the quality of research work in their field, and it is a cornerstone of scientific discovery~\cite{hettich2006mining,kelly2014peer,price2017computational}. 
In 2015 alone, approximately $63.4$ million hours were spent on peer reviews~\cite{kovanis2016global}. 
To maximize their benefit to the scientific community, it is crucial to understand and evaluate the construction and limitation of reviews themselves. However, minimal work has been done to analyze reviews' content and structure, let alone to evaluate their qualities.

\begin{figure}[t]
    \hspace{-2mm}
	\fontsize{8.5}{10.5}\selectfont
	\setlength{\tabcolsep}{0.8mm}
	\begin{tabular}{p{76mm}}
    \bottomrule
    {\bf Review $\mathbf{\#1}$} ({\it rating}: $5$, {\it \# sentences}: $11$)  \\
    \vspace{-3mm}
    {\color{gray}[}Quality: This paper demonstrates that convolutional and relational neural networks fail to solve visual relation problems $\ldots$ {\color{gray}$]_{\textsc{Fact}}$} 
    {\color{gray}[}This points at important limitations of current neural network architectures where architectures depend mainly on rote memorization.{\color{gray}$]_{\textsc{Eval}}$} $\ldots$
   {\color{gray}[}Significance: This work demonstrates failures of relational networks on relational tasks$\ldots${\color{gray}$]_{\textsc{Fact}}$} 
    {\color{gray}[}Pros: Important message about network limitations.{\color{gray}$]_{\textsc{Eval}}$}  
    {\color{gray}[}Cons: Straightforward testing of network performance on specific visual relation tasks.{\color{gray}$]_{\textsc{Eval}}$} $\ldots$ \\
	{\bf Review $\mathbf{\#2}$} ({\it rating}: $5$, {\it \# sentences}: $10$)  \\
    \vspace{-3mm}
	{\color{gray}[}The authors present two autoregressive models $\ldots${\color{gray}$]_{\textsc{Fact}}$}$\ldots$ 
    {\color{gray}[}In that context , this work can be viewed as applying deep autoregressive density estimators to policy gradient methods.{\color{gray}$]_{\textsc{Eval}}$}$\ldots$
    {\color{gray}[}At least one of those papers ought to be cited.{\color{gray}$]_{\textsc{Req}}$} 
    {\color{gray}[}It also seems like a simple, obvious baseline is missing from their experiments $\ldots${\color{gray}$]_{\textsc{Eval}}$}$\ldots$
    {\color{gray}[}The method could even be made to capture dependencies between different actions by adding a latent probabilistic layer $\ldots${\color{gray}$]_{\textsc{Eval}}$}$\ldots$
    {\color{gray}[}A direct comparison against one of the related methods in the discussion section would help{\color{gray}$]_{\textsc{Req}}$}$\ldots$\\
	\bottomrule
	\end{tabular}
	\caption{
	Sample ICLR review excerpts.
	Propositions are annotated with types, such as \textsc{Fact} (fact), \textsc{Eval} (evaluation), and \textsc{Req} (request).
	Review \#2 contains in-depth evaluation and actionable suggestion, 
	thus is perceived to be of a higher quality.
	}
\label{fig:intro}
\vspace{-3mm}
\end{figure}

As seen in Figure~\ref{fig:intro}, peer reviews resemble arguments: they contain {\bf argumentative propositions} (henceforth propositions) that convey reviewers' interpretation and evaluation of the research. Constructive reviews, e.g., review $\#2$, often contain in-depth analysis as well as concrete suggestions. 
As a result, automatically identifying propositions and their types would be useful to understand the composition of peer reviews. 

Therefore, we propose {\it an argument mining-based approach to understand the content and structure of peer reviews}. Argument mining studies the automatic detection of argumentative components and structure within discourse~\cite{peldszus2013argument}. Specifically, argument types (e.g. evidence and reasoning) and their arrangement are indicative of argument quality~\cite{P16-1150,P17-2039}. 
%
In this work, we focus on two specific tasks: (1) \textbf{proposition segmentation}---detecting elementary argumentative discourse units that are propositions, and (2) \textbf{proposition classification}---labeling the propositions according to their types (e.g., evaluation vs. request). 

Since there was no annotated dataset for peer reviews, as part of this study, we first collect $14.2$K reviews from major machine learning (ML) and natural language processing (NLP) venues. 
We create a dataset, \textsc{\bf AMPERE} (\underline{A}rgument \underline{M}ining for \underline{PE}er \underline{RE}views), by annotating $400$ reviews with $10,386$ propositions and labeling each proposition with the type of \textsc{Evaluation}, \textsc{Request}, \textsc{Fact}, \textsc{Reference}, \textsc{Quote}, or \textsc{Non-arg}.
\footnote{Dataset and annotation guideline can be found at \url{http://xinyuhua.github.io/Resources/naacl19/}.}
Significant inter-annotator agreement is achieved for proposition segmentation (Cohen's $\kappa=0.93$), with good consensus level for type annotation (Krippendorf's $\alpha_U=0.61$).

We benchmark our new dataset with state-of-the-art and popular argument mining models to better understand the challenges posed in this new domain. We observe a significant drop of performance for proposition segmentation on AMPERE, mainly due to its different argument structure. For instance, $25\%$ of the sentences contain more than one proposition, compared to that of $8\%$ for essays~\cite{J17-3005}, motivating new solutions for segmentation and classification.

We further investigate review structure difference across venues based on proposition usage, and uncover several patterns. For instance, ACL reviews tend to contain more propositions than those in ML venues, especially with more requests but fewer facts. We further find that reviews with extreme ratings, i.e., strong reject or accept, tend to be shorter and make much fewer requests. 
Moreover, we probe the salient words for different proposition types. For example, ACL reviewers ask for more ``examples" when making requests, while ICLR reviews contain more evaluation of ``network" and how models are ``trained".

\section{\textsc{AMPERE} Dataset}
\label{sec:data}
We collect review data from three sources: 
(1) \url{openreview.net}---an online peer reviewing platform 
for ICLR 2017, ICLR 2018, and UAI 2018~\footnote{ICLR reviews are downloaded from the public API: \url{https://github.com/iesl/openreview-py}. UAI reviews are collected by the OpenReview team.}; 
(2) reviews released for accepted papers at NeurIPS 
from 2013 to 2017; 
and (3) opted-in reviews for ACL 2017 from \newcite{N18-1149}. 

In total, $14,202$ reviews are collected (ICLR: $4,057$; UAI: $718$; ACL: $275$; and NeurIPS: $9,152$).
All venues except NeurIPS have paper rating scores attached to the reviews.

\smallskip
\noindent \textbf{Annotation Process.} 
For {proposition segmentation}, we adopt the concepts from~\newcite{park2015toward} and instruct the annotators to identify elementary argumentative discourse units on sentence or sub-sentence level, based on their discourse functions and topics. 
They then {classify the propositions} into five types with an additional non-argument category, as explained in Table~\ref{tab:anno-sample}. 



\begin{table}[t]
\hspace{-2mm}
\fontsize{8.5}{10.5}\selectfont
\begin{tabular}{p{74mm}}
 \hline
 {\bf \textsc{Evaluation}}: Subjective statements, often containing qualitative judgment. 
  Ex: \emph{``This paper shows nice results on a number of small tasks.''} \\
 {\bf \textsc{Request}}: Statements suggesting a course of action. 
     Ex: \emph{``The authors should compare with the following methods.''} \\
 {\bf \textsc{Fact}}: Objective information of the paper or commonsense knowledge. 
    Ex: \emph{``Existing works on multi-task neural networks typically use hand-tuned weights$\ldots$''}\\
 {\bf \textsc{Reference}}: Citations and URLs. 
     Ex: \emph{``see MuseGAN (Dong et al), MidiNet (Yang et al), etc ''} \\
 {\bf \textsc{Quote}}: Quotations from the paper. 
    Ex: \emph{``The author wrote `where r is lower bound of feature norm'.''}\\
 {\bf \textsc{Non-arg}}: Non-argumentative statements. 
    Ex: \emph{``Aha, now I understand.''}\\
 \hline
 \end{tabular}
 \vspace{-2mm}
\caption{\fontsize{10}{12}\selectfont 
Proposition types and examples. }
\label{tab:anno-sample}
\end{table}

\begin{table}[t]
\fontsize{9}{11}\selectfont
    \hspace{-2mm}
    \setlength{\tabcolsep}{0.9mm}
    \begin{tabular}{lccc}
    \bottomrule
    \textbf{Dataset} & \textbf{\#Doc} & \textbf{\#Sent} & \textbf{\#Prop} \\
    \hline
    Comments~\cite{L18-1257} & 731 & 3,994 & 4,931 \\
    Essays~\cite{J17-3005} & 402 & 7,116 & 6,089 \\
    News~\cite{C16-1324} & 300 & 11,754 & 14,313 \\
    Web~\cite{J17-1004} & 340 & 3,899 & 1,882 \\
    \rowcolor{red!20}
    AMPERE & 400 & 8,030 & 10,386 \\
   \bottomrule
    \end{tabular}
    \vspace{-2mm}
    \caption{
    Statistics for AMPERE and some argument mining corpora, including \# of annotated propositions.
    }
    \label{tab:comp-dataset}
    \vspace{-4mm}
\end{table}

    
    
    
    
    

$400$ ICLR 2018 reviews are sampled for annotation, with similar distributions of length and rating to those of the full dataset. 
Two annotators who are fluent English speakers first label the $400$ reviews with proposition segments and types, and a third annotator then resolves disagreements. 

We calculate the inter-annotator agreement between the two annotators. A Cohen's $\kappa$ of $0.93$ is achieved for proposition segmentation, with each review treated as a BIO sequence. For classification, unitized Krippendorf's $\alpha_U$~\cite{krippendorff2004measuring}, which considers disagreements among segmentation, is calculated per review and then averaged over all samples, and the value is $0.61$. Among the exactly matched proposition segments, we report a Cohen's $\kappa$ of $0.64$.


\smallskip 
\noindent \textbf{Statistics.} Table~\ref{tab:comp-dataset} shows comparison between AMPERE and some other argument mining datasets of different genres. 
We also show the number of propositions in each category in Table \ref{tab:ampere-type}.
The most frequent types are evaluation ($38.3\%$) and fact ($36.5\%$).


\begin{table}[ht]
\fontsize{9}{11}\selectfont
    \centering
    \setlength{\tabcolsep}{1.0mm}
    \begin{tabular}{ccccccc}
    \bottomrule
    \textsc{Eval} & \textsc{Req} & \textsc{Fact} & \textsc{Ref} & \textsc{Quot} & \textsc{Non-A} & Total \\
    \hline
    3,982 & 1,911 & 3,786 & 207 & 161 & 339 & 10,386 \\
    \bottomrule
    \end{tabular}
    \vspace{-1mm}
    \caption{
    Number of propositions per type in AMPERE.
    }
    \label{tab:ampere-type}
\end{table}

\vspace{-2mm}
\section{Experiments with Existing Models}
\label{sec:experiment}
We benchmark AMPERE with popular and state-of-the-art models for proposition segmentation and classification. 
Both tasks can be treated as sequence tagging problems with the setup similar to \newcite{N18-2006}. 
For experiments, $320$ reviews ($7,999$ propositions) are used for training and $80$ reviews ($2,387$ propositions) are used for testing. Following \newcite{P17-1091}, $5$-fold cross validation on the training set is used for hyperparameter tuning. To improve the accuracy of tokenization, we manually replace mathematical formulas, variables, URL links, and formatted citation with special tokens such as \texttt{<EQN>}, \texttt{<VAR>}, \texttt{<URL>}, and \texttt{<CIT>}. 
Parameters, lexicons, and features used for the models are described in the supplementary material.



\subsection{Task I: Proposition Segmentation}
\label{sec:segmentation}

We consider three baselines. 
\textbf{FullSent}: treating each sentence as a proposition. 
\textbf{PDTB-conn}: further segmenting sentences when any discourse connective (collected from Penn Discourse Treebank~\cite{prasad2007penn}) is observed. 
\textbf{RST-parser}: segmenting discourse units by the RST parser in~\newcite{P14-1048}. 
 
For learning-based methods, we start with Conditional Random Field ({\bf CRF})~\cite{lafferty2001conditional} with features proposed by \citeauthor{J17-3005} (\shortcite{J17-3005}, Table 7), and \textbf{BiLSTM-CRF}, a bidirectional Long Short-Term Memory network (BiLSTM) connected to a CRF output layer and further enhanced with ELMo representation~\cite{N18-1202}. We adopt the \textsc{BIO} scheme for sequential tagging~\cite{ramshaw1999text}, with \textsc{O} corresponding to \textsc{Non-arg}. 
Finally, we consider {\bf jointly modeling} segmentation and classification by appending the proposition types to \textsc{BI} tags, e.g., \textsc{B}-fact, with CRF (\textbf{CRF-joint}) and BiLSTM-CRF (\textbf{BiLSTM-CRF-joint}). 

Table~\ref{tab:boundary-results} shows that BiLSTM-CRF outperforms other methods in F1. 
More importantly, the performance on reviews is lower than those reached on existing datasets, e.g., an F1 of $86.7$ is obtained by CRF for essays~\cite{J17-3005}. This is mostly due to essays' better structure, with frequent use of discourse connectives.

\begin{table}[t]
\fontsize{9}{11}\selectfont
    \centering
    \setlength{\tabcolsep}{2.0mm}
    \begin{tabular}{lp{10mm}ll}
    \toprule
    
        & \textbf{Prec.} & \textbf{~Rec.} & \textbf{~~F1}  \\
    \hline
        FullSent & 73.68  & 56.00 & 63.64  \\
        PDTB-conn & 51.11  & 49.71 & 50.40   \\ 
        RST-parser & 30.28 & 43.00 & 35.54 \\ \hdashline
       
        CRF & 66.53 & 52.92 & 58.95 \\
        BiLSTM-CRF & {\bf 82.25} & {\bf 79.96} & {\bf 81.09}$^\ast$ \\
        CRF-joint & 74.99  & 63.33 & 68.67 \\
        BiLSTM-CRF-joint & 81.12  & 78.42  & 79.75 \\
    \bottomrule
    \end{tabular}
    \vspace{-1mm}
    \caption{\fontsize{10}{12}\selectfont 
    Proposition segmentation results. 
    Result that is significantly better than all comparisons is marked with $\ast$ ($p < 10^{-6}$, McNemar test). 
    }
    \label{tab:boundary-results}
\end{table}

\begin{table}[t]
\fontsize{9}{11}\selectfont
\hspace{-2mm}
\setlength{\tabcolsep}{0.6mm}
    \begin{tabular}{p{20mm}llllll}
    \toprule
         & Overall & \textsc{Eval} & \textsc{Req} & \textsc{Fact} & \textsc{Ref} & \textsc{Quot} \\
        \hline
        \multicolumn{7}{l}{\it With Gold-Standard Segments} \\
        Majority  &  40.75 & 57.90 & -- & -- & -- & -- \\
        PropLexicon &  36.83 & 40.42 & 36.07 & 32.23 & 59.57 & 31.28 \\ \hdashline
        SVM &  60.98 & 63.88 & {\bf 69.02} & 54.74 & {\bf 69.47} & 7.69 \\
        CNN &  {\bf 66.56}$^\ast$ & {\bf 69.02} & 63.26 & {\bf 66.17} & 67.44 & {\bf 52.94} \\
        \midrule
        \multicolumn{7}{l}{\it With Predicted Segments} \\
        Majority & 33.30 & 47.60 & -- & -- & -- & --\\
        PropLexicon & 23.21 & 22.45 & 23.97 & 23.73 & 35.96 & 16.67 \\ \hdashline
        SVM & 51.46 & 54.05 & 48.16 & 52.77 & 52.27 & 4.71 \\
        CNN & 55.48 & 57.75 & 53.71 & 55.19 & 48.78 & 33.33 \\
        CRF-joint &  50.69 & 46.78 & 55.74 & 52.27 & {\bf 55.77} & 26.47 \\
        BiLSTM-CRF-joint  &  {\bf 62.64}$^\ast$ & {\bf 62.36}$^\ast$ & {\bf 67.31}$^\ast$ & {\bf 61.86} & 54.74 & {\bf 37.36}  \\
    
    \bottomrule
    \end{tabular}
    \vspace{-1mm}
     \caption{\fontsize{10}{12}\selectfont 
    Proposition classification F1 scores. 
    Results that are significant better than other methods are marked with $\ast$ ($p < 10^{-6}$, McNemar test). 
    }
    \label{tab:classification-results}
\end{table}

\subsection{Task II: Proposition Classification}
\label{sec:classification}

With given proposition segments, predicted or gold-standard, we experiment with proposition-level models to label proposition types. 

We utilize two baselines. \textbf{Majority} simply assigns the majority type in the training set. \textbf{PropLexicon} matches the following lexicons for different proposition types in order, and returns the first corresponding type with a match; if no lexicon is matched, the proposition is labeled as \textsc{Non-arg}: 

\vspace{-2mm}
{\fontsize{9}{11}\selectfont
\begin{itemize}
    \item \textsc{Reference}: \texttt{<URL>}, \texttt{<CIT>} \vspace{-3mm}
    \item \textsc{Quote}: ``, '', ' \vspace{-3mm}
    \item \textsc{Request}:  \textit{should, would be nice, why, please, would like to, need} \vspace{-3mm}
    \item \textsc{Evaluation}:  \textit{highly, very, unclear, clear, interesting, novel, well, important, similar, clearly, quite, good} \vspace{-3mm}
    \item  \textsc{Fact}:  \textit{author, authors, propose, present, method, parameters, example, dataset, same, incorrect, correct}
\end{itemize}
}



For supervised models, we employ linear \textbf{SVM} with a squared hinge loss and group Lasso regularizer~\cite{yuan2006model}. It is trained with the top $500$ features selected from Table 9 in~\cite{J17-3005} by $\chi^2$ test. We also train a convolutional neural network (\textbf{CNN}) proposed by~\newcite{D14-1181}, with the same setup and pre-trained word embeddings from word2vec~\cite{mikolov2013distributed}. Finally, results by joint models of CRF and BiLSTM-CRF are also reported. 
%

F1 scores for all propositions and each type are reported in Table~\ref{tab:classification-results}. A prediction is correct when both segment and type are matched with the true labels. 
CNN performs better for types with significantly more training samples, i.e., evaluation and fact, indicating the effect of data size on neural model's performance. 
Joint models (CRF-joint and BiLSTM-CRF-joint) yield the best F1 scores for all categories when gold-standard segmentation is unavailable. 

\section{Proposition Analysis by Venues}
\label{sec:analysis}
Here we leverage the BiLSTM-CRF-joint model trained on the annotated AMPERE data to identify propositions and their types in unlabeled reviews from the four venues (ICLR, UAI, ACL, and NeurIPS), to understand the content and structure of peer reviews at a larger scale. 

\begin{figure}[h]
\subfloat
{	
	\hspace{-4mm}
    \includegraphics[width=40mm]{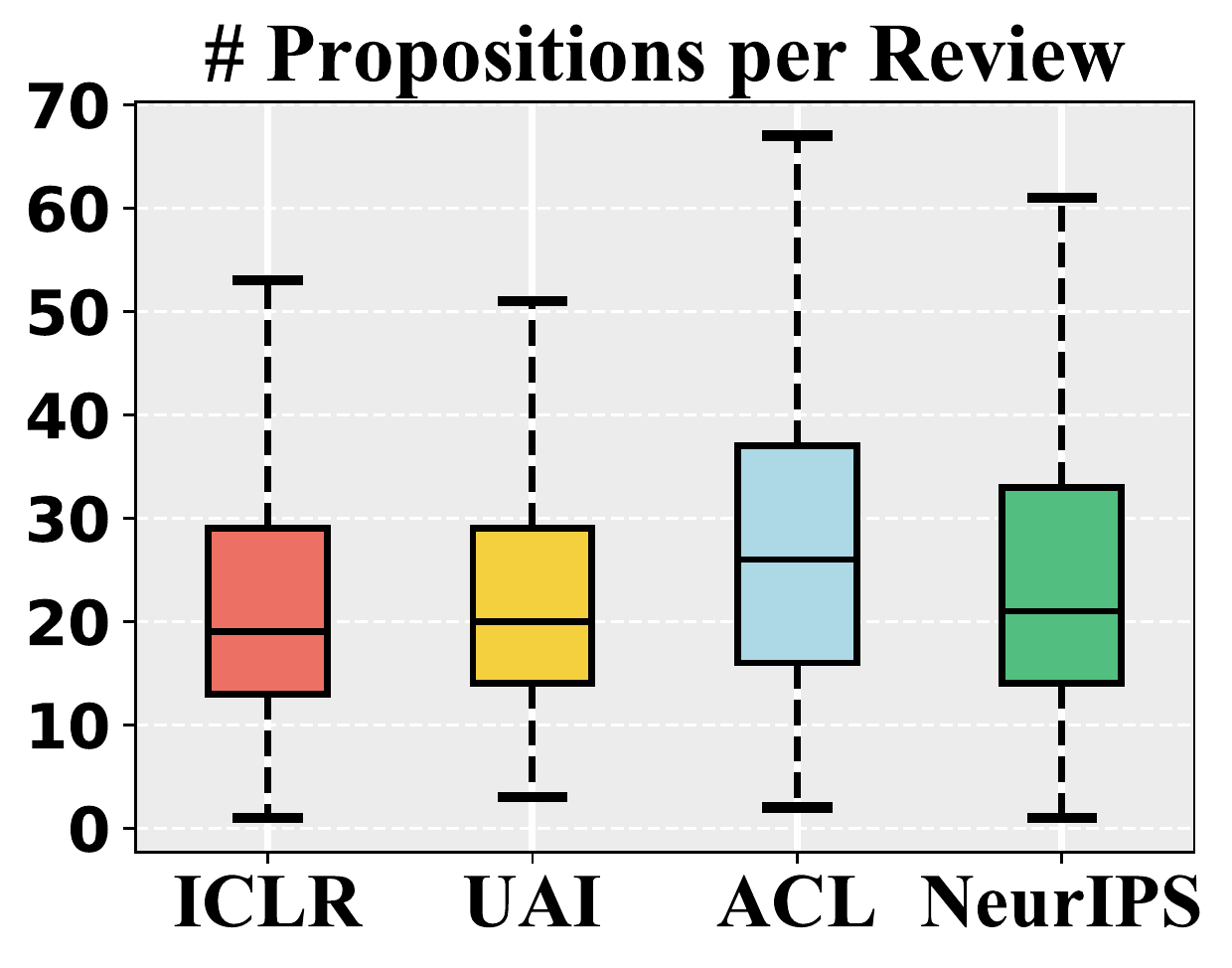}
}
\subfloat
{	
	\hspace{-3mm}
    \includegraphics[width=40mm]{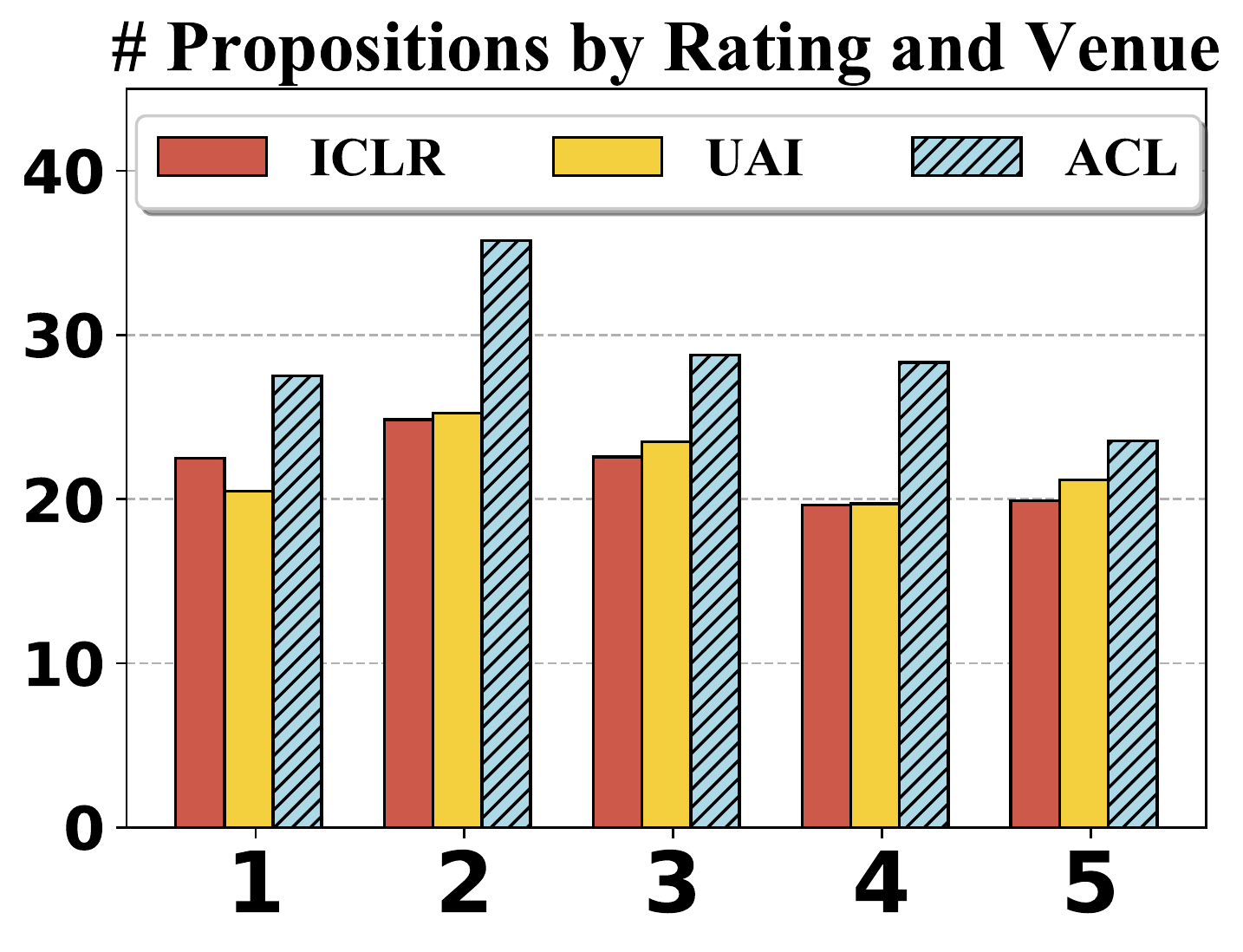}
}
\vspace{-3mm}
 \caption{
 Proposition number in reviews. Differences among venues are all significant except UAI vs. ICLR and ACL vs. NeurIPS ($p < 10^{-6}$, unpaired $t$-test). 
  }
\label{fig:length-dist}
\end{figure}

\smallskip
\noindent \textbf{Proposition Usage by Venue and Rating.} 
Figure \ref{fig:length-dist} shows the {average} number of propositions per review, grouped by venue and rating.  
Scores in $1-10$ are scaled to $1-5$ by $\lceil x/2\rceil$, with $1$ as strong reject and $5$ as strong accept.
 ACL and NeurIPS have significantly more propositions than ICLR and UAI. 
Ratings, which reflect a reviewer's judgment of paper quality, also affect proposition usage. 
We find that reviews with extreme ratings, i.e., $1$ and $5$, tend to have fewer propositions. 

\begin{figure}[ht]
\centering
\hspace{-2.3mm}
    \includegraphics[width=78mm]{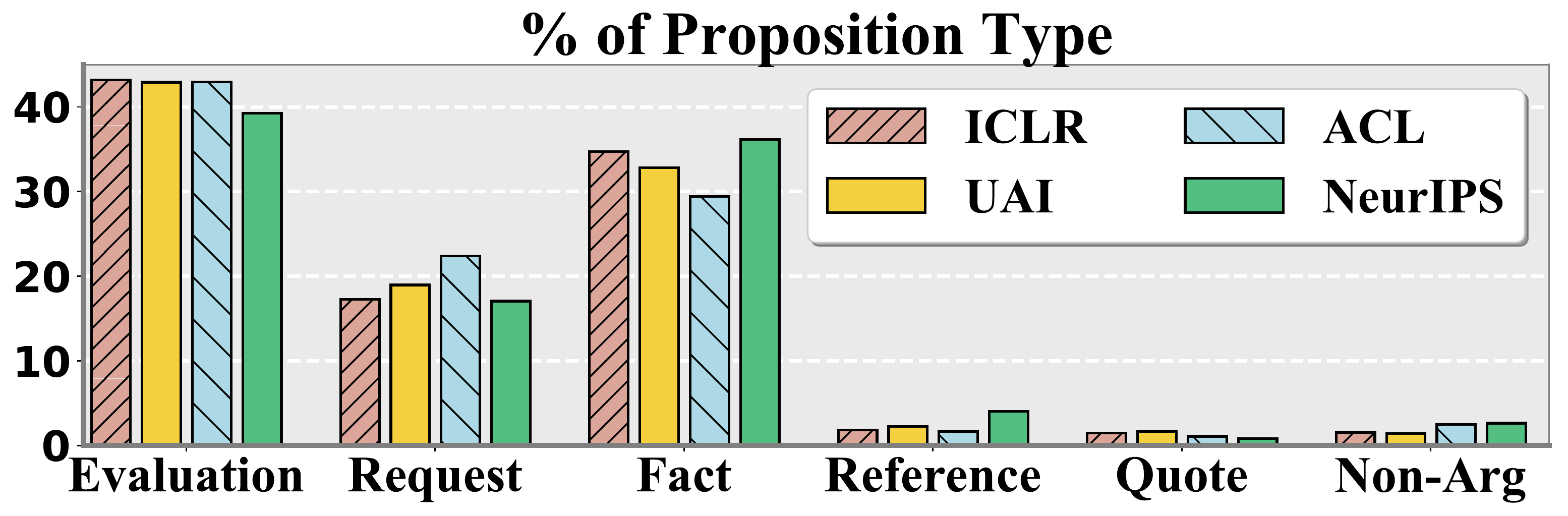}
     \vspace{-6mm}
    \caption{
    Distribution of proposition type per venue. 
    }
    \label{fig:type-venue}
\end{figure}

\begin{figure}[ht]
\centering
    \includegraphics[width=72mm]{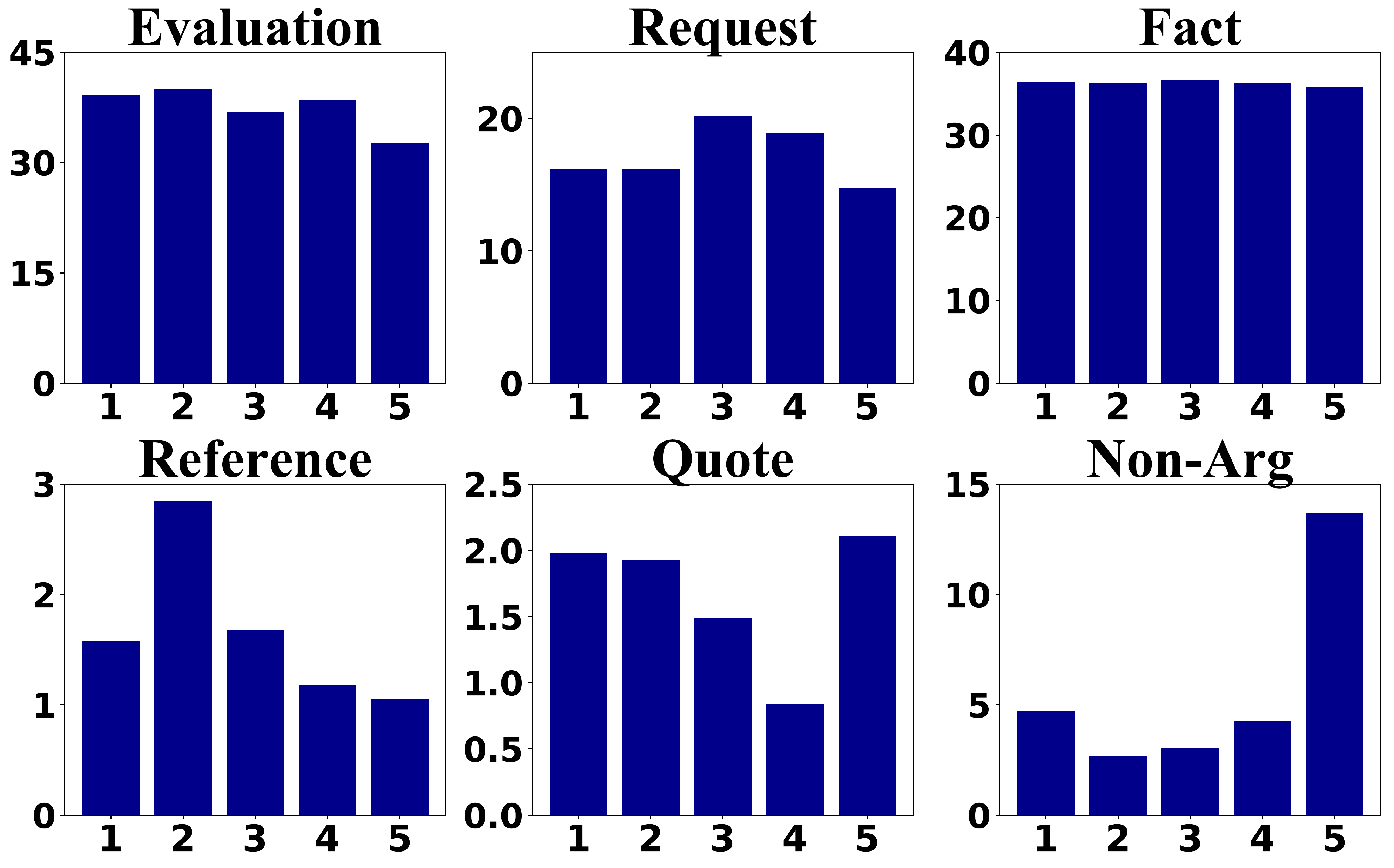}
    \vspace{-4mm}
    \caption{
    Distribution of proposition type per rating (in $\%$) on AMPERE. 
    }
    \label{fig:type-rating}
\end{figure}

We further study the distribution of proposition type in each venue. 
As observed in Figure~\ref{fig:type-venue}, ACL reviews contain more {request}s but fewer {fact}s than other venues. Specifically, we find that $94.6\%$ of ACL reviews have at least one \textsc{request} proposition, compared to $81.5\%$ for ICLR and $84.7\%$ for UAI. 
We also show proposition type distribution based on ratings in Figure~\ref{fig:type-rating}. Reviews with the highest rating tend to use fewer evaluation and reference, while reviews with ratings of $3-4$ (borderline or weak accept) contain more requests. We further observe a sharp decrease of \textsc{Quote} usage in rating group $4$, and a surge of \textsc{Non-arg} for rating group $5$, while \textsc{Fact} remains consistent across rating ranges.

\begin{table*}[t]
\fontsize{8.5}{10}\selectfont
\setlength{\tabcolsep}{0.8mm}
    \centering
    \begin{tabular}{|l|p{27mm}|p{27mm}|p{27mm}|p{27mm}|p{27mm}|}
    \hline
         & \textsc{Evaluation} & \textsc{Request} & \textsc{Fact} & \textsc{Reference} & \textsc{Quote} \\
         \hline\hline
        {\bf All Venues} &   overall, unclear, not, contribution, seem, interesting & please, could, should, if, why, would, more, suggest &  think, each, some, data, useful, written, proposes & \texttt{<URL>}, et, al., conference, paper, proceedings, arxiv & ", paper, we, :, our \\
        \hline
        {\bf ICLR} & network, general, acceptance, convinced, trained & network, appendix, recommend, because, novelty & training, results, work, then, image & deep, ;, nips, pp., speech & not, section, 4, 5, agent  \\
        \hline
        {\bf UAI} & quality, relevant, found, presentation, major & \texttt{<VAR>}, model, method, nice, column & stochastic, called, considers, sense, writing & artificial, discovery, etc., via, systems & --, second, column, processes, connections \\
        \hline
        {\bf ACL} & weaknesses, strengths, so, word, main & consider, examples, further, models, proposed & word, method, words, proposed, embeddings & language, extraction, emnlp, computational, linguistics & \\
        \hline
        {\bf NeurIPS} & theoretical, \texttt{<EQN>}, interest, practical, nips & following, clarity, address, significance, quality & \texttt{<EQN>}, maximum, may, comments, characters & for, see, class, detailed, guidelines & of, in, which, \texttt{<EQN>}, reviewer \\
        \hline
    \end{tabular}
    \vspace{-1mm}
    \caption{
    Salient words ($\alpha=0.001$, $\chi^2$ test) per proposition type. 
    Top $5$ frequent words that are unique for each venue are shown. ``\texttt{<EQN>}", ``\texttt{<URL>}'', and ``\texttt{<VAR>}" are equations, URL links, and variables. 
    }
    \label{tab:salient-words}
\end{table*}

\smallskip
\noindent \textbf{Proposition Structure.} 
Argumentative structure, which is usually studied as support and attack relations, reveals how propositions are organized into coherent text. 
According to~\newcite{L18-1257}, $75\%$ of support relations happen between adjacent propositions in user comments. We thus plot the proposition transition probability matrix in Figure~\ref{tab:type-transit}, to show the argument structure in AMPERE.
The high probabilities along the diagonal line imply that propositions of the same type are often constructed consecutively,
with the exception of {quote}, which is more likely to be followed by {evaluation}. 

\begin{figure}[t]
\fontsize{9}{9.5}\selectfont
    \centering
     \setlength{\tabcolsep}{0.5mm}
    \begin{tabular}{|l|cccccc|}
    \hline
      & \textsc{Eval} & \textsc{Req} &  \textsc{Fact} & \textsc{Ref} & \textsc{Quot} & \textsc{Non-A} \\
    \hline
    \textsc{Eval} & \cellcolor{red!50}50.3 & \cellcolor{red!17}17.2 & \cellcolor{red!27}27.3 & \cellcolor{red!1}1.0 & \cellcolor{red!1}1.4 & \cellcolor{red!3}2.9 \\
    \textsc{Req} & \cellcolor{red!32}32.2 & \cellcolor{red!42}41.6 & \cellcolor{red!19}19.4 & \cellcolor{red!2}1.8 & \cellcolor{red!2}2.3 & \cellcolor{red!3}2.8 \\
    \textsc{Fact} & \cellcolor{red!34}33.5 & \cellcolor{red!11}11.0 & \cellcolor{red!51}51.2 & \cellcolor{red!1}1.3 & \cellcolor{red!1}0.9 & \cellcolor{red!2}2.0 \\
    \textsc{Ref} & \cellcolor{red!15}15.0 & \cellcolor{red!11}10.8 & \cellcolor{red!18}18.0 & \cellcolor{red!51}50.9 & \cellcolor{red!4}3.6 & \cellcolor{red!2}1.8 \\
    \textsc{Quot} & \cellcolor{red!31}31.2 & \cellcolor{red!24}23.6 & \cellcolor{red!26}25.5 & \cellcolor{red!1}1.3 & \cellcolor{red!12}12.1 & \cellcolor{red!6}6.4 \\
    \textsc{Non-A} & \cellcolor{red!32}31.9 & \cellcolor{red!16}15.5 & \cellcolor{red!23}22.7 & \cellcolor{red!1}1.3 & \cellcolor{red!3}2.8 & \cellcolor{red!26}25.9 \\
   \hline
    \end{tabular}
    \vspace{-1mm}
    \caption{
   Proposition transition prob. on AMPERE. 
    }
    \label{tab:type-transit}
\end{figure}

\smallskip
\noindent \textbf{Proposition Type and Content.}
We also probe the salient words used for each proposition type, and the difference of their usage across venues. For each venue, we utilize log-likelihood ratio test~\cite{C00-1072} to identify the representative words in each proposition type compared to other types. 
Table~\ref{tab:salient-words} shows both the commonly used salient words across venues and the unique words with top frequencies for each venue ($\alpha=0.001$, $\chi^2$ test). 
For {evaluation}, all venues tend to focus on clarity and contribution, with ICLR discussing more about ``network" and NeurIPS often mentioning equations. ACL reviews then frequently request for ``examples". 

\section{Related Work}
\label{sec:related}
There is a growing interest in understanding the content and assessing the quality of peer reviews. 
Authors' feedback such as satisfaction and helpfulness have been adopted as quality indicators~\cite{latu2000review,hart2010method,P11-2088}. 
Nonetheless, they suffer from author subjectivity and are often influenced by acceptance decisions~\cite{weber2002author}. 
Evaluation by experts or editors proves to be more reliable and informative~\cite{van1999development}, but requires substantial work and knowledge of the field. 
Shallow linguistic features, e.g., sentiment words, are studied in~\newcite{bornmann2012closed} for analyzing languages in peer reviews. 
To the best of our knowledge, our work is the first to understand the content and structure of peer reviews via argument usage.

Our work is also in line with the growing body of research in argument mining~\cite{teufel1999argumentative,palau2009argumentation}. 
Most of the work focuses on arguments in social media posts~\cite{W14-2105, P16-2032, P16-1150}, online debate portals or Oxford-style debates~\cite{P17-2039, hua-wang:2017:Short, Q17-1016}, and student essays~\cite{P15-1053, P16-2089}. 
We study a new domain of peer reviews, and identify new challenges for existing models.

\section{Conclusion}
\label{sec:conclusion}
We study the content and structure of peer reviews under the argument mining framework. 
AMPERE, a new dataset of peer reviews, is collected and annotated with propositions and their types. 
We benchmark AMPERE with state-of-the-art argument mining models for proposition segmentation and classification.  
We leverage the classifiers to analyze the proposition usage in reviews across ML and NLP venues, showing interesting patterns in proposition types and content.


\section*{Acknowledgements}
This research is based upon work supported in part by National Science Foundation through Grants
IIS-1566382 and IIS-1813341. 
We are grateful to the OpenReview team, especially Michael Spector, for setting up the API to facilitate review data collection. 
We also thank three anonymous reviewers for their constructive suggestions. 

\bibliography{ref}
\bibliographystyle{acl_natbib}

\appendix
\section{Annotation Details}
\label{sec:anno}
\paragraph{Data Selection.}
We select $400$ reviews from the ICLR 2018 dataset for the annotation study. To ensure the subset is representative of the full dataset, samples are drawn based on two aspects: review length and rating score. 

Table \ref{tab:len-dist} shows the distribution of reviews with regard to their length in the full ICLR 2018 dataset and the subset we sampled for annotation (AMPERE). 
As can be seen, the distribution over five bins are consistent between AMPERE and full dataset. A similar trend is observed on rating distribution in Table \ref{tab:rat-dist}.

A subset of the reviews also have revision history, which can be used as a proxy for opinion change and review quality in future work. To that end, we manually set the ratio of revised reviews vs. unrevised ones to $3$:$1$ (c.f. $9$:$1$ on the full ICLR2018 dataset), to ensure that enough revised reviews are being annotated. Notice that, in this study, we only consider the initial version of a review if any revision exists.

\begin{table}[ht]
 \centering
    \fontsize{9}{11}\selectfont
    \setlength{\tabcolsep}{0.5mm}
    \begin{tabular}{p{13mm}lllll}
    \toprule 
    {\bf Length} & (0,200] & (200,400] & (400,600] & (600,800] & (800,$\infty$)  \\
    \hline
    AMPERE & 14.8$\%$ & 35.5$\%$ & 25.3$\%$ & 10.0$\%$ &14.6$\%$ \\
    ICLR2018 & 17.6$\%$ & 39.3$\%$ & 23.8$\%$ & 11.4$\%$ & 7.9$\%$  \\
    \bottomrule
    \end{tabular}
    \caption{ 
    Review length distribution of the full ICLR 2018 dataset and AMPERE, which consists of 400 sampled reviews. 
    }\label{tab:len-dist}
\end{table}

\begin{table}[ht]
 \centering
    \fontsize{9}{11}\selectfont
    \setlength{\tabcolsep}{0.8mm}
    \begin{tabular}{p{15mm}lllll}
    \toprule 
    {\bf Rating} & {\bf 1} & {\bf 2} & {\bf 3} & {\bf 4} & {\bf 5} \\
    \hline
    AMPERE &  3.0$\%$ & 32.5$\%$ & 43.8$\%$ & 19.3$\%$ & 1.5$\%$ \\
    ICLR2018 & 2.6$\%$ & 32.5$\%$ & 42.4$\%$ & 20.6$\%$ & 1.8$\%$ \\
    \bottomrule
    \end{tabular}
    \caption{ 
    Review rating distribution of AMPERE and the full ICLR 2018 dataset.
    }\label{tab:rat-dist}
\end{table}

\paragraph{Inter-annotator Agreement (IAA).}
To measure IAA, we first follow \newcite{J17-3005} to calculate the unitized Krippendorf's $\alpha_U$~\cite{krippendorff2004measuring} for each review, and report the average for each type. 

We further consider agreement on the proposition level. However, since the segmented proposition boundaries by two annotators do not always match, we only consider the exact matched segments for Cohen's $\kappa$. The agreement scores for each type are listed in Table \ref{tab:iaa}.

\begin{table}[ht]
\fontsize{9}{11}\selectfont
    \centering
    \setlength{\tabcolsep}{1.2mm}
    \begin{tabular}{llllllll}
    \bottomrule
    & \textsc{Eval} & \textsc{Req} & \textsc{Fact} & \textsc{Ref} & \textsc{Quot} & \textsc{Non-A} & overall \\
    \hline
    \bf{$\alpha_U$} & 0.51 & 0.64 & 0.60 & 0.63 & 0.41 & 0.18 & 0.61 \\
    \bf{$\kappa$} & 0.60 & 0.68 & 0.64 & 0.88 & 0.59 & 0.27 & 0.64 \\
 \bottomrule
    \end{tabular}
    \vspace{-2mm}
    \caption{\fontsize{10}{12}\selectfont 
    Inter-annotator agreement for all categories.
    }
    \label{tab:iaa}
\end{table}

\paragraph{Sample Annotations.} We show examples of annotated propositions in Table \ref{tab:sample}.

\begin{table}[h]
    \centering
    \fontsize{8}{10}\selectfont
    \setlength{\tabcolsep}{0.2mm}
    \begin{tabular}{|p{5mm}|p{72mm}|}
    \hline
    \parbox[t]{2mm}{\multirow{8}{*}{\rotatebox[origin=c]{90}{\textsc{Evaluation}}}} & The paper shows nice results on a number of small tasks. \\
    \cline{2-2}
    &  With its poor exposition of the technique, it is difficult to recommend this paper for publication. \\\cline{2-2}
    &  I like the general approach of explicitly putting desired equivariance in the convolutional networks. \\\cline{2-2}
    &  The paper covers a very interesting topic and presents some though-provoking ideas. \\\cline{2-2}
    &  I'm not sure this strong language can be justified here. \\
    \hline
     \parbox[t]{2mm}{\multirow{6}{*}{\rotatebox[origin=c]{90}{\textsc{Request}}}} & I would really like to see how the method performs without this hack.\\\cline{2-2}
    & can the authors motivate this aspect better? \\\cline{2-2}
    &  I suggest using [hidelinks] for hyperref.\\\cline{2-2}
    & More explanation needed here. \\\cline{2-2}
    & In addtion -$>$ In addition \\
    \hline
     \parbox[t]{2mm}{\multirow{12}{*}{\rotatebox[origin=c]{90}{\textsc{Fact}}}} & Existing works on multi-task neural networks typically use hand-tuned weights for weighing losses across different tasks \\\cline{2-2}
    & This work proposes a dynamic weight update scheme that updates weights for different task losses during training time by making use of the loss ratios of different tasks \\\cline{2-2}
    &  In this paper, the authors trains a large number of MNIST classifier networks with differing attributes (batch-size, activation function, no. layers etc.)\\\cline{2-2}
    & This paper is based on the theory of group equivariant CNNs (G-CNNs), proposed by Cohen and Welling ICML'16. \\
    \hline
     \parbox[t]{2mm}{\multirow{8}{*}{\rotatebox[origin=c]{90}{\textsc{Reference}}}} & [1] Burnetas, A. N., \& Katehakis, M. N. (1997). Optimal adaptive policies for Markov decision processes. Mathematics of Operations Research, 22(1) , 222-255 \\\cline{2-2}
    & VARIANCE-BASED GRADIENT COMPRESSION FOR EFFICIENT DISTRIBUTED DEEP
LEARNING \\\cline{2-2}
    & see MuseGAN (Dong et al), MidiNet (Yang et al), etc \\\cline{2-2}
    &  e.g. “Weakly-supervised Disentangling with Recurrent Transformations for 3D View
Synthesis”, Yang et al. \\
    \hline
    \parbox[t]{2mm}{\multirow{7}{*}{\rotatebox[origin=c]{90}{\textsc{Quote}}}} & The author wrote ``where r is lower bound of feature norm'' \\\cline{2-2}
    & ``In a probabilistic context-free grammar (PCFG), all production rules are independent'' \\\cline{2-2}
    &  Quoting from its abstract: ``Using commodity hardware, our implementation achieves $\sim$
90\% scaling efficiency when moving from 8 to 256 GPUs.''\\
    \hline
    \parbox[t]{2mm}{\multirow{4}{*}{\rotatebox[origin=c]{90}{\textsc{Non-Arg}}}} & Did I miss something here? \\\cline{2-2}
    & Below, I give some examples \\\cline{2-2}
    &  are all the test images resized before hand?\\\cline{2-2}
    & How was this chosen?
    \\
    \hline
    \end{tabular}
    \caption{Sample annotated propositions.}
    \label{tab:sample}
\end{table}

\section{Experiments}
\label{sec:exp}
\subsection{Data Preprocessing} 
For preprocessing, we tokenize and split reviews into sentences with the Stanford CoreNLP toolkit~\cite{manning-EtAl:2014:P14-5}. 
We manually substitute special tokens for mathematical equations, URLs, and citations or references. 
In total, $302$ variables (\texttt{<VAR>}), $125$ equations (\texttt{<EQN>}), $62$ URL links (\texttt{<URL>}), and $97$ citations (\texttt{<CIT>}) are identified in $400$ reviews. 

\subsection{Training Details}
For all models except CNN, we conduct $5$-fold cross validation on training set to select hyperparameters.

 \paragraph{CRF.} We utilize the CRFSuite~\cite{CRFsuite} implementation and tune coefficients $C_1$ and $C_2$ for $\ell_1$ and $\ell_2$ regularizer. For segmentation task the optimal setup is $C_1=0.0$ and $C_2=1.0$; for joint prediction, $C_1=1.0$ and $C_2=0.01$ is used. 
 
 \paragraph{BiLSTM-CRF.} We experiment with implementation by \newcite{reimers-gurevych:2017:EMNLP2017} with an extra ELMo embedding. Based on the cross validation for both segmentation and joint learning, the optimal network architecture selected has two layers with $100$ dimensional hidden states each, with dropout probabilities of $0.5$ for both layers. 
 The word embedding pre-trained by \newcite{N16-1175} is chosen, as it outperforms GloVe embeddings~\cite{D14-1162} trained either on Google News or Wikipedia.
 
 \paragraph{SVM.} We utilize SAGA~\cite{defazio2014saga} implemented in the Lightning library~\cite{blondel2016lightning} to learn a linear SVM optimized with Coordinate Descent~\cite{wright2015coordinate}. The coefficient for a group Lasso regularizer~\cite{yuan2006model} is set to $0.001$ by cross validation.
 
 \paragraph{CNN.} We implement the CNN-non-static variant as described in \newcite{D14-1181}, with the following configuration: filter window sizes of $\{3$,$4$,$5\}$, with $128$ feature maps each. Dropout probability is $0.5$. $300$ dimensional word embeddings are initiated with the pre-trained word2vec on $100$ billion Google News~\cite{mikolov2013distributed}.

\section{Further Analysis}
\label{sec:analysis-app}
\begin{figure}[t]
\centering
\subfloat
{	
    \includegraphics[width=55mm]{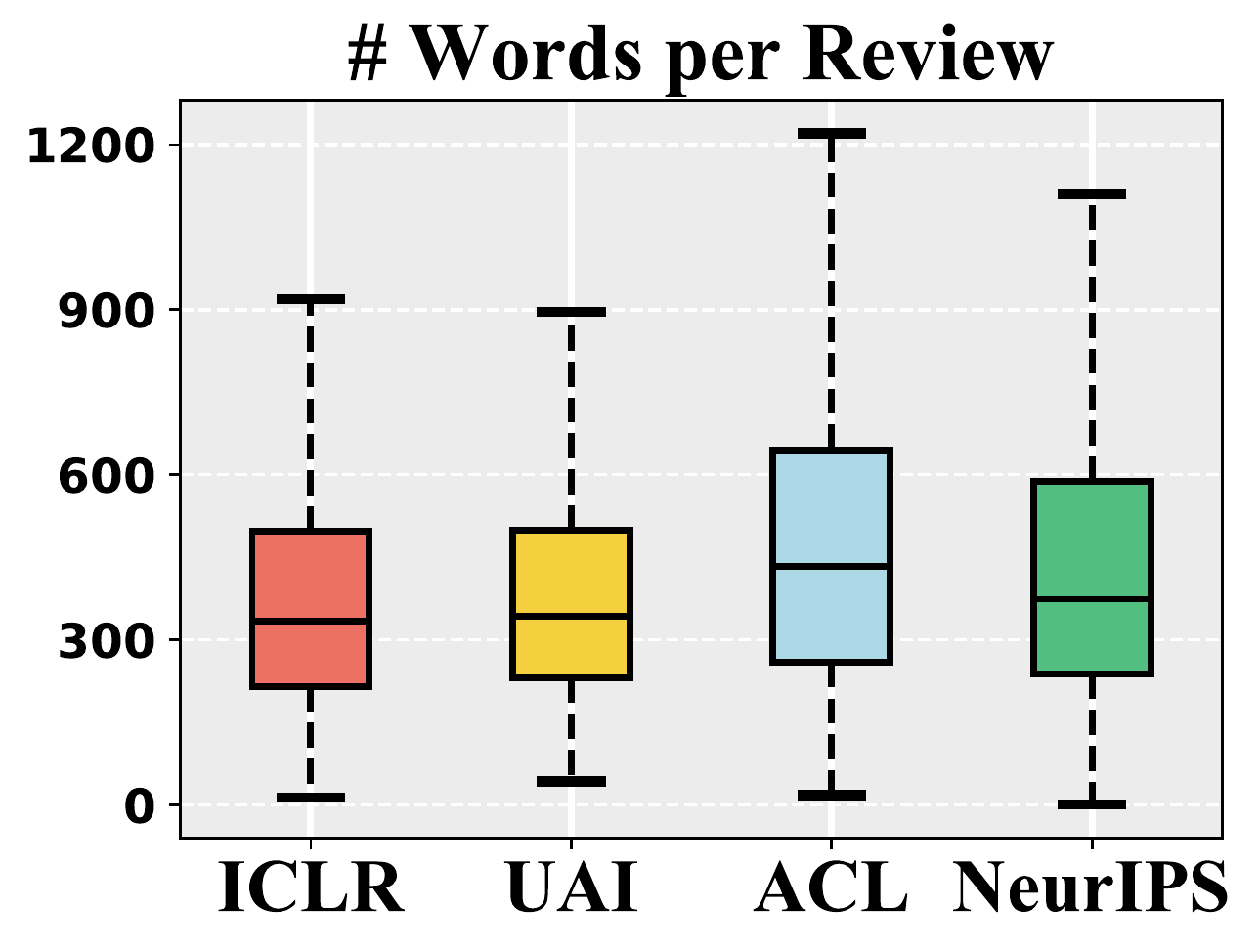}
}\\
\subfloat
{	
    \includegraphics[width=55mm]{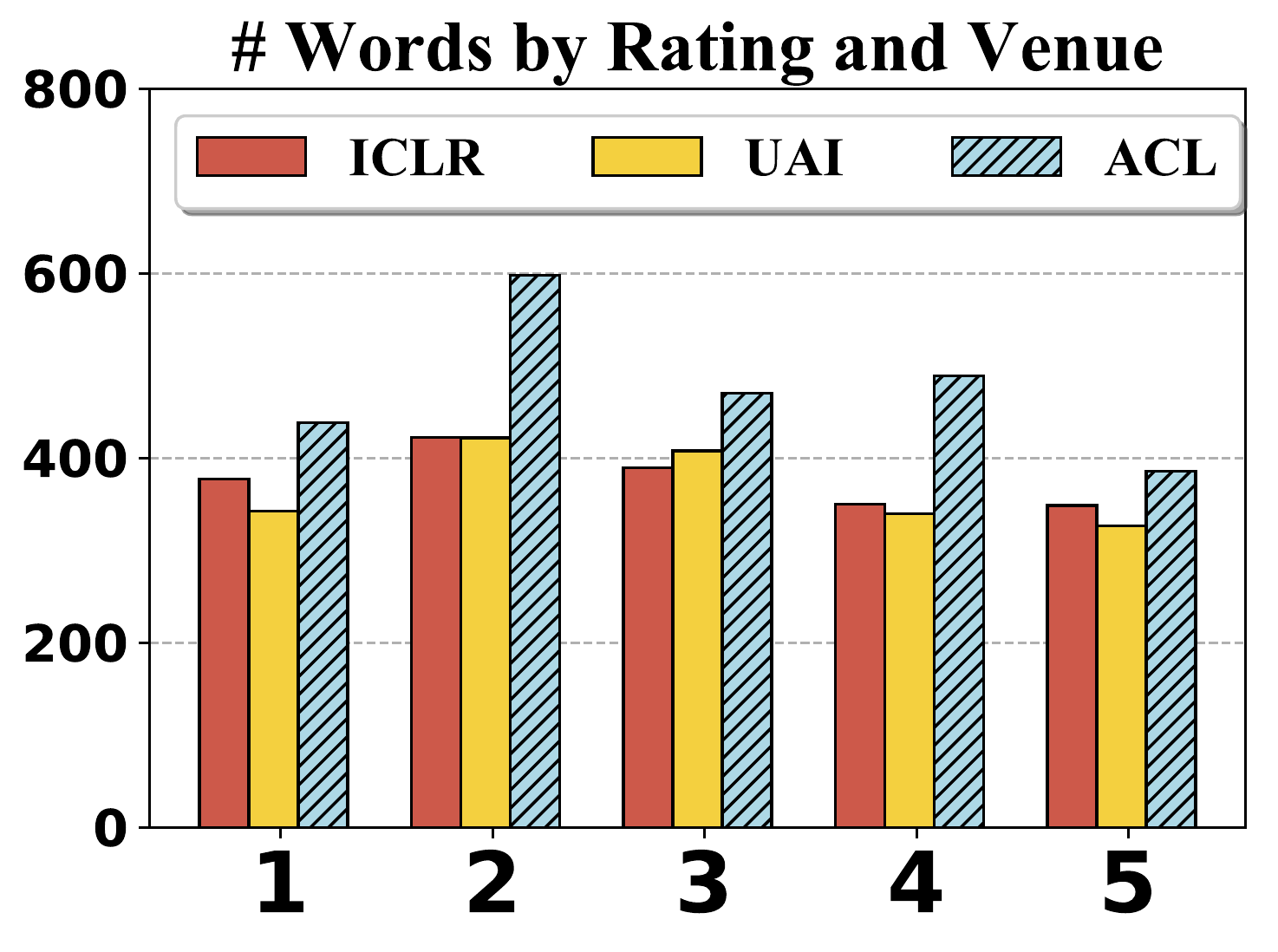}
}
 \caption{
 Word count in reviews by venue and rating. The word counts are significantly different between all venue pairs except UAI vs. ICLR and ACL vs. NeurIPS ($p < 10^{-6}$, unpaired $t$-test).}
\label{fig:length-dist-words}
\end{figure}

\paragraph{Review Length by Venue and Rating.}
We compare review length of different venues in the top row of Figure \ref{fig:length-dist-words}. Unpaired $t$-test shows that ACL and NeurIPS have significantly longer reviews than UAI and ICLR ($p < 10^{-6}$), which is consistent with the trend for proposition counts, as described in Figure 2 in the paper.

We further group reviews by their ratings and display the average length per category in Figure \ref{fig:length-dist-words}. Again, we observe similar trends for the distribution of proposition count, where reviews with extreme ratings tend to be shorter. 

\begin{figure}[t]
\subfloat
{	
	\hspace{-3mm}
    \includegraphics[width=40mm]{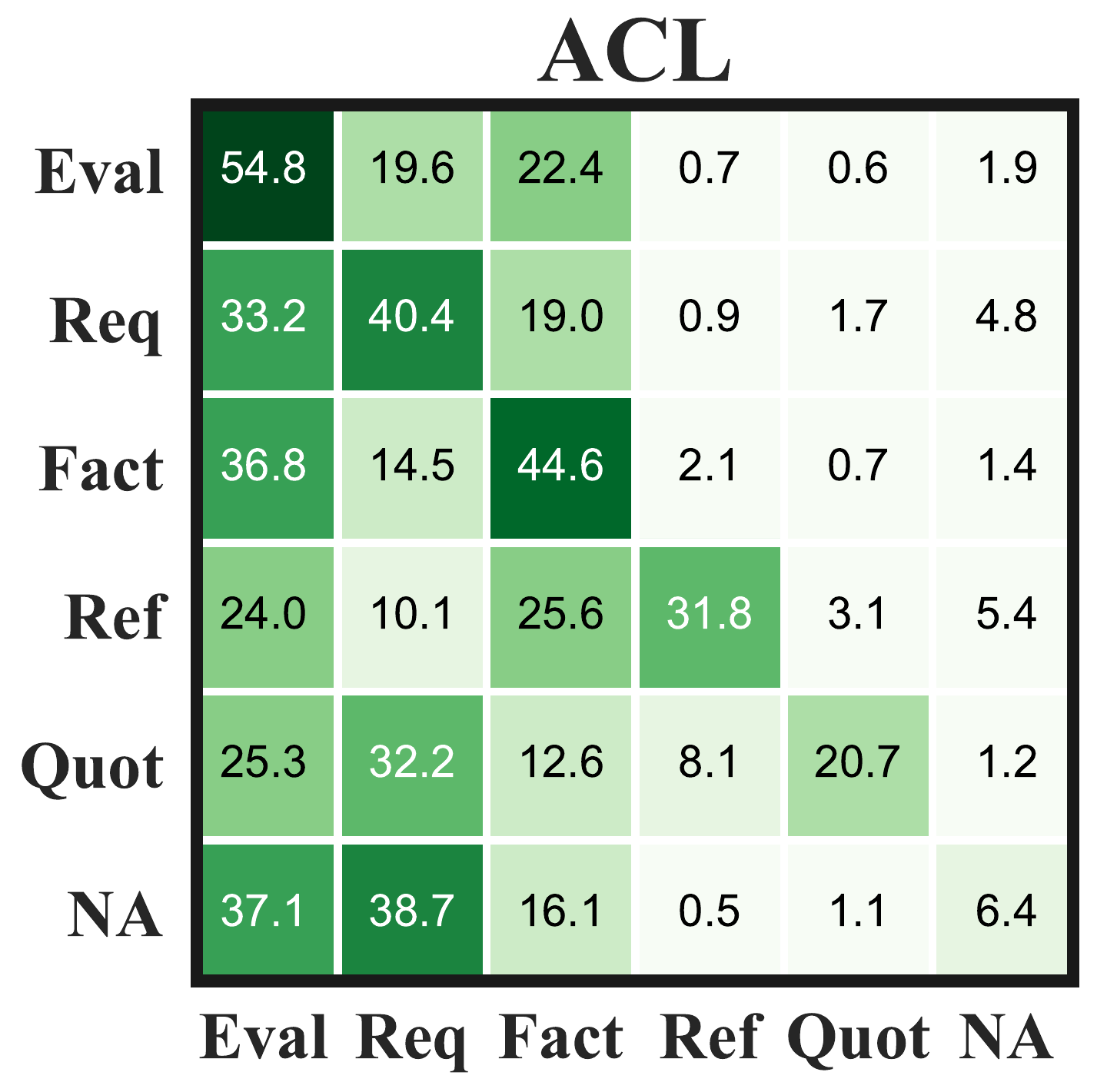}
}
\subfloat
{	
	\hspace{-4mm}
    \includegraphics[width=40mm]{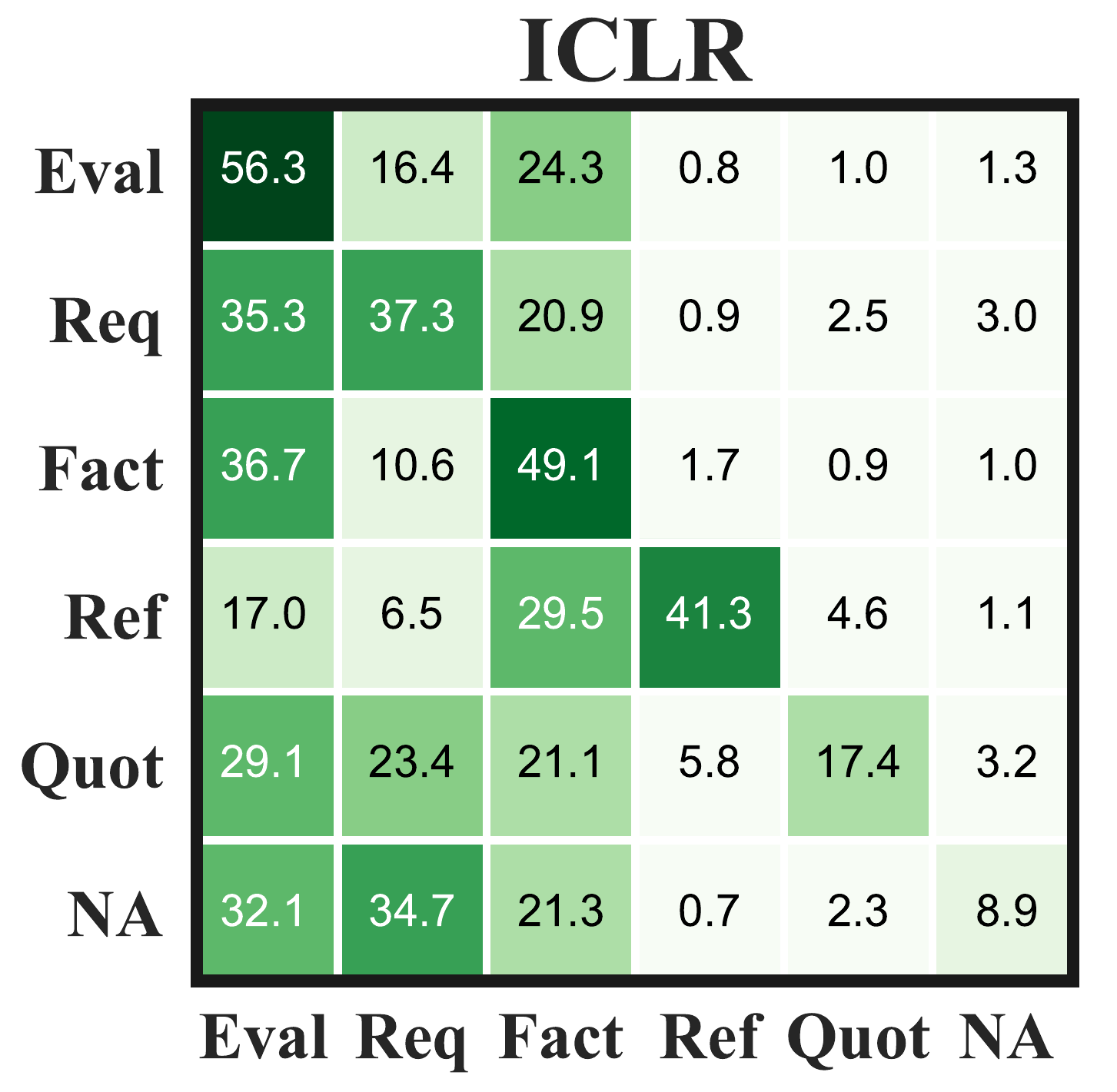}
}\\
\subfloat
{	
	\hspace{-3mm}
    \includegraphics[width=40mm]{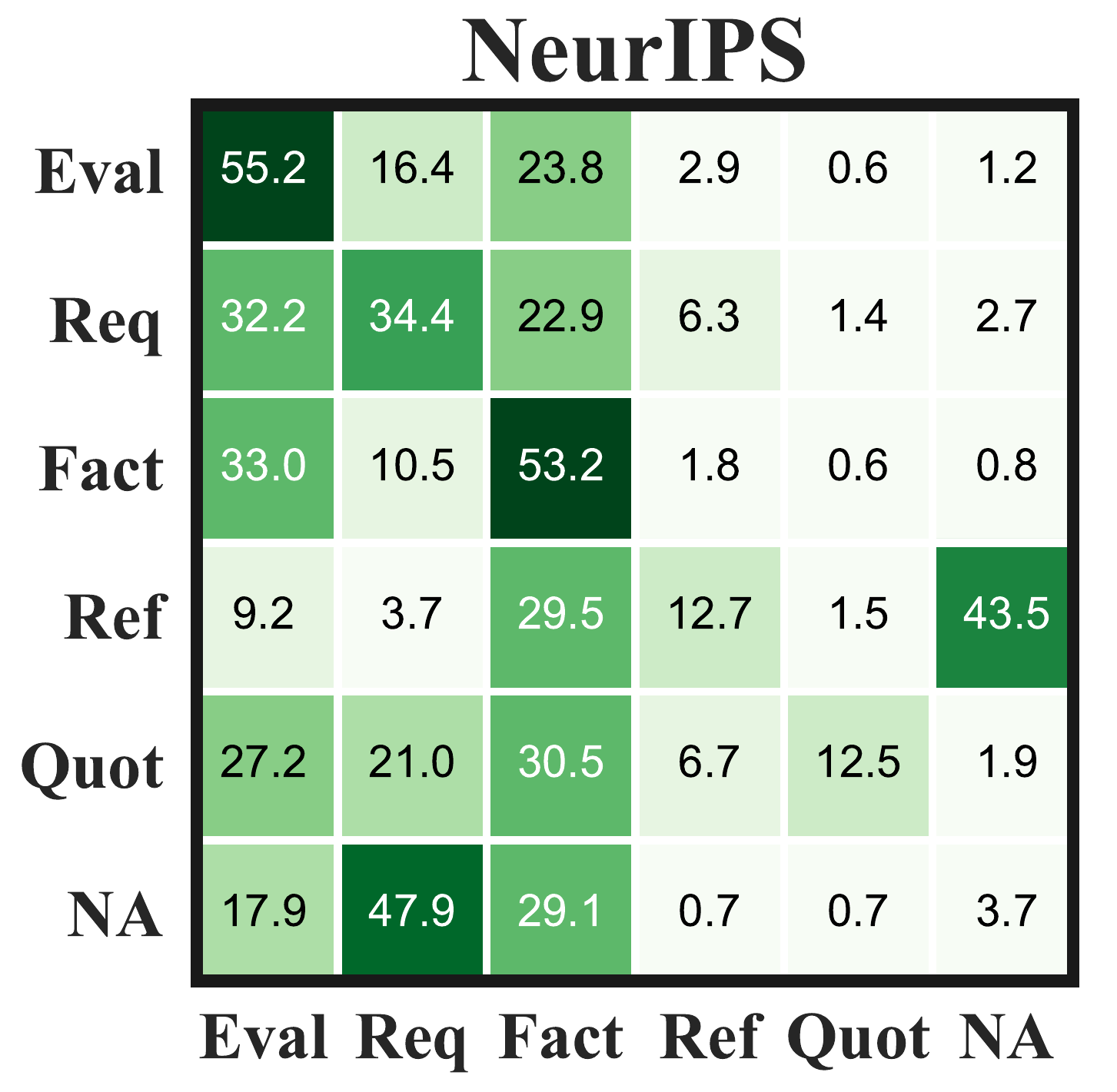}
}
\subfloat
{	
	\hspace{-4mm}
    \includegraphics[width=40mm]{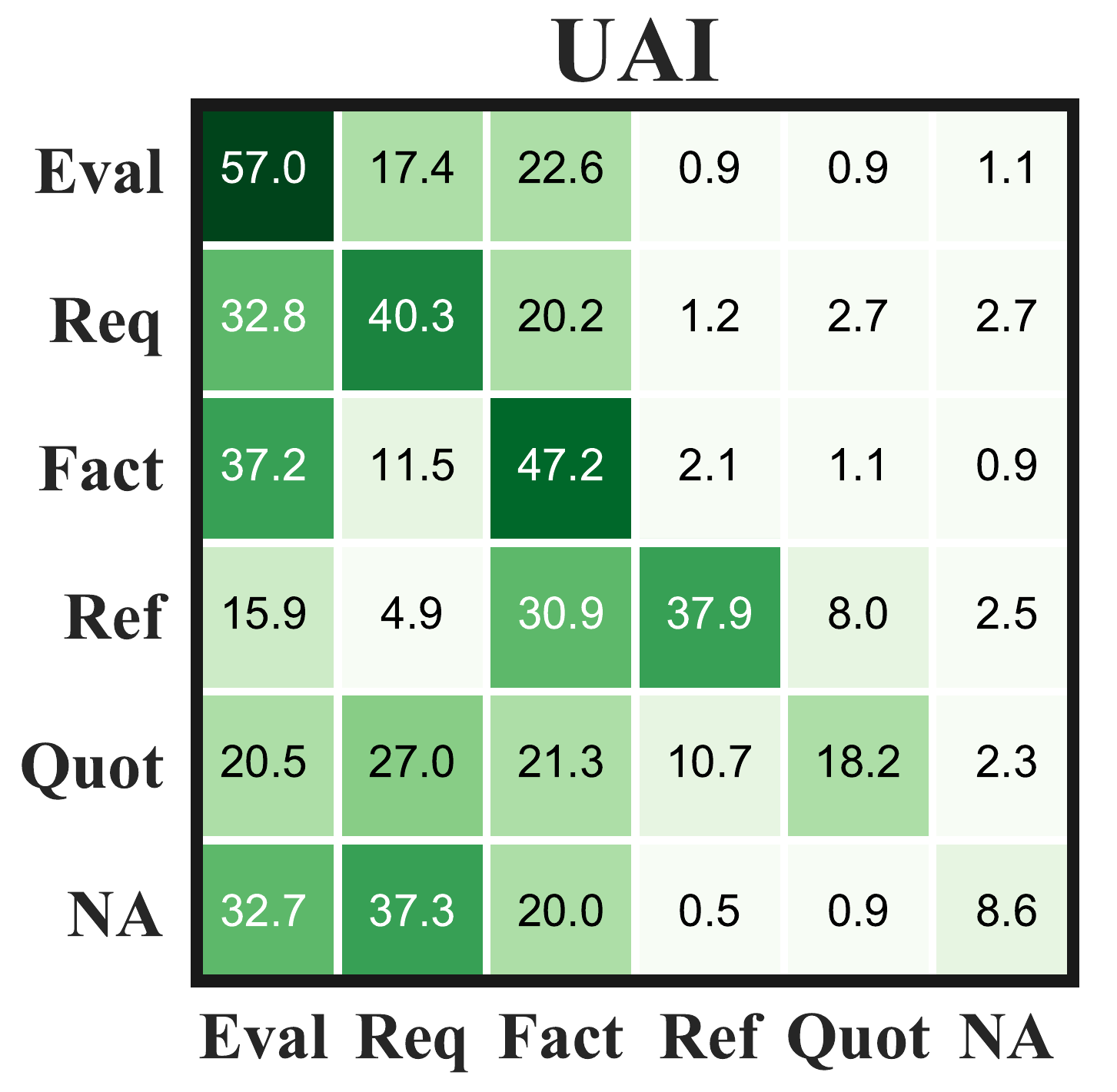}
}
 \caption{ 
 Proposition type transition matrix in different venues.}
\label{fig:transit}
\end{figure}

\paragraph{Proposition Structure.}
We calculate the proposition type transition matrix as a proxy to uncover the local argumentative structure information.  As is shown in Figure \ref{fig:transit}, propositions are more likely to be followed by propositions of the same type, while for NeurIPS the transition from reference to non-argument is much more prominent than other venues. 
A closer look at the dataset indicates that this might be because many formatted headers are mistakenly predicted as reference, e.g. ``For detailed reviewing guidelines, see \texttt{<URL>}''. They are usually followed by text such as ``Comments to the author'', which is predicted correctly as \textsc{Non-arg}.

\end{document}